%% file: acl_latex.tex
\pdfoutput=1
\PassOptionsToPackage{dvipsnames}{xcolor} % \textcolor clash 해결
\documentclass[11pt]{article}

\usepackage[]{acl}

\usepackage[most]{tcolorbox}
\usepackage{times}
\usepackage{latexsym}
\usepackage[utf8]{inputenc}
\usepackage[T1]{fontenc}
\usepackage{CJKutf8}
\usepackage[english]{babel}
\usepackage{makecell}
\usepackage{graphicx}
\usepackage{hyperref}
\usepackage{multirow} % Table 기능
\usepackage{float} % H
\usepackage{amsmath} % cases / align
\usepackage{adjustbox} % \resizebox{\textwidth}{!}
\usepackage{verbatim} % comment 기능
\usepackage{hwemoji}
\usepackage{pifont} % emoji 
\usepackage[normalem]{ulem}
\useunder{\uline}{\ul}{}
\usepackage[table,xcdraw]{xcolor}  % \textcolor 확장
\usepackage{array}
\usepackage[T1]{fontenc}
\usepackage[utf8]{inputenc}

\usepackage{microtype}

\newcommand\blfootnote[1]{%
  \begingroup
  \renewcommand\thefootnote{}\footnote{#1}%
  \addtocounter{footnote}{-1}%
  \endgroup
}

\title{LLM Agents at the Roundtable: A Multi-Perspective and Dialectical Reasoning Framework for Essay Scoring}

\author{
  Jinhee Jang$^{1,2}$ \quad
  Ayoung Moon$^{1}$ \quad
  Minkyoung Jung$^{1}$ \quad
  YoungBin Kim$^{2}$\footnotemark[2] \quad
  Seung Jin Lee$^{1}$\footnotemark[2] \\
  $^1$NC AI \\
  $^2$Chung-Ang University \\
  \texttt{\{evelynm, jmk95, sjlee927\}@ncsoft.com}, \\
  \texttt{\{jinheejang, ybkim85\}@cau.ac.kr}
}

\begin{document}
\maketitle

\blfootnote{\textsuperscript{†}Corresponding authors.}

\begin{abstract}
The emergence of large language models (LLMs) has brought a new paradigm to automated essay scoring (AES), a long-standing and practical application of natural language processing in education. However, achieving human-level multi-perspective understanding and judgment remains a challenge. In this work, we propose Roundtable Essay Scoring (RES), a multi-agent evaluation framework designed to perform precise and human-aligned scoring under a zero-shot setting. RES constructs evaluator agents based on LLMs, each tailored to a specific prompt and topic context. Each agent independently generates a trait-based rubric and conducts a multi-perspective evaluation. Then, by simulating a roundtable-style discussion, RES consolidates individual evaluations through a dialectical reasoning process to produce a final holistic score that more closely aligns with human evaluation.
By enabling collaboration and consensus among agents with diverse evaluation perspectives, RES outperforms prior zero-shot AES approaches. Experiments on the ASAP dataset using ChatGPT and Claude show that RES achieves up to a 34.86\% improvement in average QWK over straightforward prompting (Vanilla) methods.
\end{abstract}

\begin{figure*}[t]
    \centering
    \includegraphics[width=\textwidth]{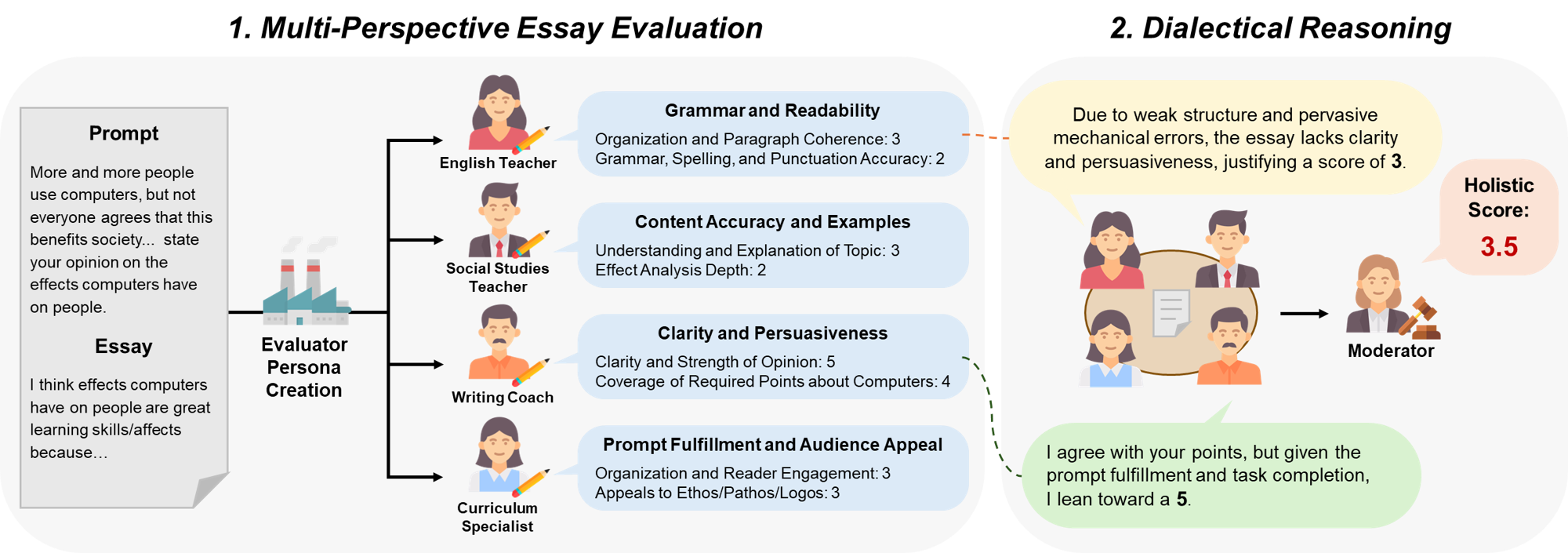}
    \caption{The RES framework overview illustrates a two-stage process in which multi-perspective LLM agents evaluate an essay and collaboratively derive a holistic score through dialectical reasoning.}
\label{fig:framework}
\end{figure*}

\section{Introduction}
\label{sec:intro}
\input{01_introduction}

\input{tables/table1.tex}

\section{Roundtable Essay Scoring (RES)}
\label{sec:res}
\input{02_method}

\section{Experiment}
\label{sec:experiment}
\input{03_experiments}

\section{Conclusion}
\label{sec:conclusion}
\input{04_conclusion}

\section*{Limitations}
While the proposed RES framework demonstrates notable improvements in zero-shot essay evaluation, several limitations of our current approach warrant further exploration.

First, RES is currently implemented using API-based proprietary language models, which may incur higher usage costs compared to locally deployed open-source alternatives. This choice was made to ensure access to state-of-the-art model performance and to reflect realistic application scenarios in production environments. Moreover, the framework requires no fine-tuning or large-scale infrastructure and can be executed with minimal computational resources, making it relatively accessible and reproducible even for researchers and practitioners without high-performance computing resources. We have also conducted additional experiments on open-source LLMs (e.g., Qwen3-4B, Qwen3-8B) within the limits of our available resources. However, we found that these models exhibited weaker instruction-following capabilities compared to high-performance commercial API-based models, which led to challenges in interpreting and parsing their outputs when using prompts designed for more capable models. While prior studies on LLM-as-evaluator \cite{leiter2024prexme, kartac2025open} offer indirect insights into the capabilities of open-source LLMs, applying the RES framework to such models in the context of essay evaluation remains a relevant and practical direction for future work.

Second, although one of the key strengths of LLMs lies in their capacity not only to assign scores but also to generate informative, rationale-based feedback, this study focuses exclusively on scoring. While prior works have examined feedback generation in various domains \cite{liu2023geval, du2024llms, koutcheme2025evaluating}, the reliability and pedagogical value of LLM-generated feedback in essay evaluation—especially at the content and discourse level—remain underexplored \cite{stahl2024exploring}. As a next step, we plan to expand the current multi-agent framework to evaluate and enhance feedback quality, aiming to address not only grammar but also higher-level aspects such as content and structure.

\section*{Ethics Statement}
We used only publicly available, anonymized essay datasets and did not include any personally identifiable or sensitive information. The language models employed in our experiments were accessed via official APIs. We acknowledge that commercial models are trained on non-public data, which may contain inherent biases.
This research was conducted solely for academic purposes, and we emphasize that applying the system to educational decision-making without human oversight is not appropriate \cite{li2024llms, fabiano2024ai}.

\section*{Acknowledgements}
This work was partly supported by the Institute of Information \& Communications Technology Planning \& Evaluation (IITP) grant funded by the Korea government(MSIT) (No. RS-2025-25441313, Professional AI Talent Development Program for Multimodal AI Agents, Contribution: 50\%) and the Institute of Information \& Communications Technology Planning \& Evaluation (IITP) grant funded by the Korea government (MSIT) [No. RS-2021-II211341, Artificial Intelligence Graduate School Program (Chung-Ang University), Contribution: 25\%] and by the National Research Foundation of Korea (NRF) grant funded by the Korea government (MSIT) (No. RS-2025-00556246, Contribution: 25\%).
% This work was supported by the Institute of Information \& Communications Technology Planning \& Evaluation (IITP) grant funded by the Korea government (MSIT) [RS-2021-II211341, Artificial Intelligence Graduate School Program (Chung-Ang University)] and by the National Research Foundation of Korea (NRF) grant funded by the Korea government (MSIT) (RS-2025-00556246).
%ACK HERE


\bibliography{custom}
\bibliographystyle{acl_natbib}

\input{09_appendix}

\end{document}

%% file: 01_introduction.tex
Essay scoring has long been used in educational settings as a tool for evaluating students' reasoning, expression, and thinking skills in a comprehensive manner \cite{white1985teaching, bereiter2013psychology}. However, human evaluation is often limited by the high cost and time required \cite{dikli2006overview}. To address this issue, automated essay scoring (AES) systems have been actively studied for decades, with efforts focusing on improving efficiency and reliability through statistical features and machine learning techniques \cite{ke2019automated}.

With the development of neural networks, AES research has largely shifted toward supervised learning approaches. These include multi-trait scoring methods that assign independent scores to grammar, organization, content relevance, and other traits \cite{wang2023orthogonal, do2024autoregressive, chu2025rationale}, as well as cross-prompt essay scoring techniques aimed at generalizing across different essay prompts \cite{do2023prompt, li2024conundrums, do2025prompt}. However, supervised approaches face limitations in domain transfer and scalability without access to large-scale labeled data \cite{lee2024unleashing}.

As a result, zero-shot AES, which evaluates essays by directly prompting large language models (LLMs) without additional training, has gained increasing attention \cite{yancey2023rating, lee2024unleashing, stahl2024exploring, shibata2025lceszero}. Leveraging the vast linguistic knowledge and evaluative priors acquired during pretraining, LLMs can assess not only surface-level traits but also more complex dimensions such as topic relevance, depth of content, and contextual coherence \cite{yehudai2025survey, zhao2025survey, gu2025surveyllmasajudge}. Furthermore, the use of multiple LLMs in a collaborative, multi-agent setup has shown promise in handling higher-order evaluation tasks \cite{guo2024multiagents, chan2024chateval, chen2024llmarena, koupaee2025faithful}.

Nevertheless, most existing zero-shot AES methods remain limited to static prompting of a single LLM to predict scores, and compared to other text evaluation tasks \cite{liu2023geval, chiang2024llmeval, koupaee2025faithful}, the potential of LLMs in AES remains relatively underexplored. As a result, LLMs have thus far served merely as replacements for traditional AES models, rather than realizing their full potential for collaborative reasoning.

To address this limitation, we propose \textbf{RES} (\textbf{R}oundtable \textbf{E}ssay \textbf{S}coring), a multi-agent framework for AES. Inspired by dialectical deliberation in human assessment processes \cite{mercier2011humans}, RES involves multiple LLM-based evaluator agents—each with a distinct evaluative perspective—who independently construct rubrics and evaluate the essay. These agents then engage in a dialectical reasoning process to discuss their evaluation results and rationales, ultimately producing a holistic score.

To validate the effectiveness of the RES framework, we conduct evaluations on the ASAP essay dataset. Results show that RES improves performance by 13.19\% with ChatGPT and 34.86\% with Claude, in terms of QWK, compared to the Vanilla approach. This demonstrates that the collaborative, multi-perspective structure of RES enhances alignment with human scoring.

%% file: tables/table1.tex
\begin{table*}[t]
\centering
\small
\resizebox{0.9\textwidth}{!}{%
\begin{tabular}{c|c|cccccccc|c}
\Xhline{5\arrayrulewidth}
\textbf{Model}   & \textbf{Method}                    & \textbf{P1}                            & \textbf{P2}                            & \textbf{P3}                            & \textbf{P4}                            & \textbf{P5}                            & \textbf{P6}                            & \textbf{P7}                            & \textbf{P8}                            & \textbf{Average}                       \\ \hline\hline
\textbf{}        & Vanilla                            & 0.063                                  & 0.184                                  & 0.213                                  & 0.545                                  & 0.43                                   & 0.559                                  & 0.249                                  & 0.672                                  & 0.364                                  \\
\textbf{ChatGPT} & MTS                                & 0.157                                  & \textbf{0.357}                         & 0.328                                  & 0.607                                  & \textbf{0.522}                         & 0.592                                  & 0.337                                  & 0.401                                  & 0.412                                  \\
\textbf{}        & \cellcolor[HTML]{EFEFEF}RES (ours) & \cellcolor[HTML]{EFEFEF}\textbf{0.229} & \cellcolor[HTML]{EFEFEF}0.334          & \cellcolor[HTML]{EFEFEF}\textbf{0.415} & \cellcolor[HTML]{EFEFEF}\textbf{0.628} & \cellcolor[HTML]{EFEFEF}0.487          & \cellcolor[HTML]{EFEFEF}\textbf{0.606} & \cellcolor[HTML]{EFEFEF}\textbf{0.457} & \cellcolor[HTML]{EFEFEF}\textbf{0.713} & \cellcolor[HTML]{EFEFEF}\textbf{0.483} \\ \hline
\textbf{}        & Vanilla                            & 0.137                                  & 0.345                                  & 0.175                                  & 0.556                                  & 0.38                                   & 0.425                                  & 0.454                                  & 0.491                                  & 0.370                                  \\
\textbf{Claude}  & MTS                                & \textbf{0.174}                         & 0.303                                  & 0.107                                  & 0.333                                  & 0.344                                  & 0.402                                  & 0.258                                  & 0.511                                  & 0.304                                  \\
\textbf{}        & \cellcolor[HTML]{EFEFEF}RES (ours) & \cellcolor[HTML]{EFEFEF}0.163          & \cellcolor[HTML]{EFEFEF}\textbf{0.422} & \cellcolor[HTML]{EFEFEF}\textbf{0.323} & \cellcolor[HTML]{EFEFEF}\textbf{0.708} & \cellcolor[HTML]{EFEFEF}\textbf{0.657} & \cellcolor[HTML]{EFEFEF}\textbf{0.630} & \cellcolor[HTML]{EFEFEF}\textbf{0.500} & \cellcolor[HTML]{EFEFEF}\textbf{0.587} & \cellcolor[HTML]{EFEFEF}\textbf{0.499} \\
\Xhline{5\arrayrulewidth}
\end{tabular}
}
\caption{QWK scores for zero-shot AES performance on the ASAP dataset. P1–8 denotes Prompts 1 to 8. The highest score for each prompt is shown in bold.}
\label{tab:tab1}
\end{table*}

%% file: 02_method.tex
RES is a multi-agent framework that evaluates essays from multiple perspectives through collaborative dialectical reasoning. As shown in Figure~\ref{fig:framework}, RES operates in two stages: Multi-Perspective Evaluation and Dialectical Reasoning.

\subsection{Multi-Perspective Essay Evaluation}
The goal of the multi-perspective essay evaluation stage is to conduct a fine-grained, multi-perspective pre-evaluation in preparation for the subsequent Dialectical Reasoning stage. The instructions used in this stage are provided in Appendices~\ref{app:instruction1}, ~\ref{app:instruction2}, and ~\ref{app:instruction3}

\noindent\textbf{Evaluator Persona Creation.}
As shown on the left side of Figure ~\ref{fig:framework}, the input includes the essay prompt, the essay itself, and optionally, metadata such as the writer’s grade level or essay type. Based on this input, evaluator personas are generated to match the essay topic and student proficiency (e.g., a middle school teacher for an 8th-grade essay), preventing the use of criteria that are too lenient or too strict.

\noindent\textbf{Automated Rubric Construction.}
Each persona is assigned to an LLM-based evaluator agent, which then independently constructs a rubric based on the essay prompt and contextual information. The rubric consists of multi-traits that reflect various aspects of writing, such as grammar, organization, coherence, and topic relevance. The rubric is designed not only to reflect general essay traits but also to incorporate criteria specific to the essay’s topic and purpose.

\noindent\textbf{Rationale-first Multi-trait Evaluation.}
Each evaluator agent independently evaluates the essay based on its own rubric. Before assigning trait-level scores, the agent generates rationales that provide justifications for each score. This approach, inspired by \citet{yancey2023rating}, makes the agent’s scoring criteria and interpretive perspective explicit. Furthermore, the rationales enhance the explainability of the evaluation process and serve as a foundation for generating an appropriate final score in the subsequent dialectical reasoning stage.

\subsection{Dialectical Reasoning}
In the dialectical reasoning stage, illustrated on the right side of Figure~\ref{fig:framework}, evaluator personas engage in a simulated roundtable discussion to integrate multi-perspective results, exchange critiques, and coordinate their judgments into a final holistic score. The instruction is provided in Appendix~\ref{app:instruction4}

\noindent\textbf{Dialectical Dialogue Simulation.}
In this stage, each persona presents its rationale, and a dialectical dialogue unfolds, involving support and counterarguments. This process takes place within a single generation, allowing multiple personas to build upon each other’s reasoning in a cumulative manner, which facilitates deeper deliberation. The structure is similar in spirit to the Solo Performance Prompting (SPP) method proposed by \citet{wang2024unleashing}.

\noindent\textbf{Moderator-led Dialectical Synthesis.}
After the dialogue, a moderator agent synthesizes the discussion, resolves disagreements, and assigns the final holistic score. By analyzing each evaluator agent’s arguments, identifying points of agreement and conflict, and providing balanced justifications, the moderator ensures the validity of the outcome.

%% file: 03_experiments.tex
\subsection{Experimental Settings}

\noindent\textbf{LLM Models.}
To ensure practical applicability and align with state-of-the-art standards, we used API-based models: GPT-4.1-mini-2025-04-14\footnote{https://platform.openai.com/docs/models/gpt-4.1-mini} (\textbf{ChatGPT}) and Claude-3.5-haiku-20241022\footnote{https://www.anthropic.com/claude/haiku} (\textbf{Claude}). These models offer strong zero-shot performance without fine-tuning and are widely adopted in commercial evaluation systems, including essay scoring \cite{gu2025surveyllmasajudge}.

\noindent\textbf{Baselines.}
We compared RES with two zero-shot AES baselines. The first, \textbf{Vanilla}, improves performance by generating rationales before scoring \cite{yancey2023rating}, using the same instruction as in \citet{lee2024unleashing}. The second, \textbf{Multi-Trait Specialization (MTS)} \cite{lee2024unleashing}. uses a human-designed rubric to derive trait-specific criteria, evaluates each trait individually, and aggregates the results into a holistic score through normalization.

\noindent\textbf{Dataset.}
We used the ASAP\footnote{https://www.kaggle.com/c/asap-aes/data} (Automated Student Assessment Prize) dataset for evaluation. The dataset consists of 12,978 essays written by students from grades 7 to 10 in response to eight different prompts,  Detailed statistics of the dataset are provided in Table~\ref{tab:tab3}. Following the setup in \citet{lee2024unleashing}, we used 10\% of the test split for evaluation.

\noindent\textbf{Evaluation Metric.}
We used the Quadratic Weighted Kappa (QWK) as the main evaluation metric. QWK measures the agreement between model-predicted scores and human-assigned scores, ranging from -1 to 1. A higher value indicates greater consistency between the two sets of scores.

\subsection{Main Results}
Table~\ref{tab:tab1} shows the main results of zero-shot AES experiments on the ASAP dataset. We evaluated Vanilla, MTS, and RES using both ChatGPT and Claude models (Performance comparisons with prompt-based methods other than zero-shot AES are provided in Appendix~\ref{app:app_more_comparison}.). For RES, four evaluator agents were used, each generating and evaluating three traits, resulting in 12 trait-based criteria.

\input{tables/table2.tex}

Across most experiments with both models, RES consistently outperformed other methods. With ChatGPT, RES achieved the highest performance, averaging 0.483. MTS improved upon Vanilla (0.364), reaching 0.412, but showed instability on certain prompts (e.g., P8). A similar pattern was observed with Claude, where RES averaged 0.499, outperforming both Vanilla (0.370) and MTS (0.304).

Notably, while MTS relied on rubrics manually constructed by human raters from the ASAP dataset, RES did not use any input rubrics. Instead, each evaluator agent autonomously generated its own rubric and conducted the evaluation accordingly. Despite this, RES achieved a higher alignment with human evaluation through dialectical reasoning among multiple agents.

\subsection{Ablation Study}
We conduct an ablation study to analyze the effects of key components in the RES framework, namely the number of evaluator agents ($N_{\text{Agents}}$), the total number of evaluated traits ($N_{\text{Traits}}$), and the inclusion of Dialectical Reasoning (DR).

\begin{figure}[t]
    \centering
    \includegraphics[width=0.9\linewidth]{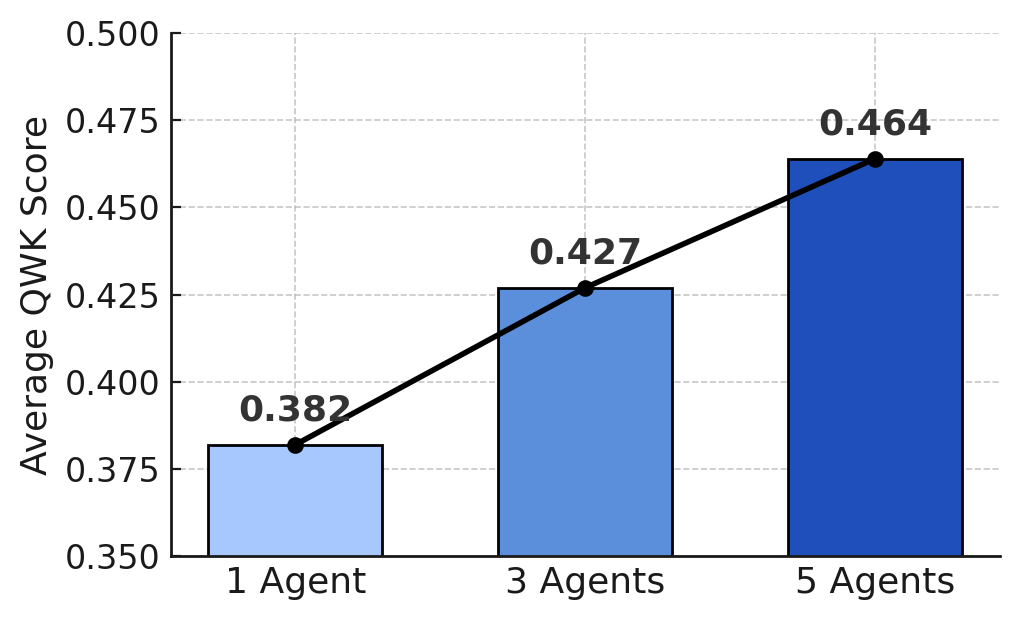}
    \caption{QWK scores by number of evaluator agents ($N_{\text{Agents}}$), averaged over eight prompts using ChatGPT.}
    \label{fig:n_agents}
\end{figure}

First, to examine the impact of the number of evaluator agents, we fixed $N_{\text{Traits}} = 3$ and varied $N_{\text{Agents}} = 1, 3, 5$. As shown in Figure~\ref{fig:n_agents}, increasing the number of agents led to a steady improvement in the average QWK across the eight prompts. Notably, increasing from 1 to 3 agents resulted in a significant improvement of approximately 11.8\%, whereas increasing from 3 to 5 agents yielded a more modest gain of 8.7\%, indicating diminishing returns. This suggests the importance of determining an optimal number of agents from an efficiency standpoint.

\begin{figure}[t]
    \centering
    \includegraphics[width=0.9\linewidth]{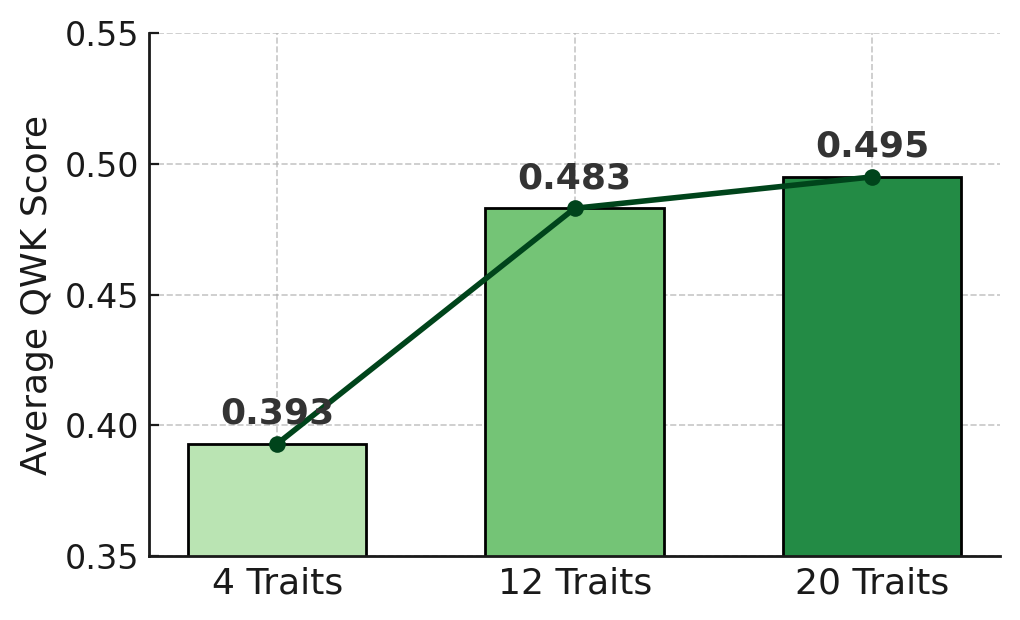}
    \caption{Average QWK scores across eight prompts using ChatGPT, with four evaluator agents. The total number of evaluation traits ($N_{\text{Traits}}$) refers to the combined traits across all agents.}
    \label{fig:n_traits}
\end{figure}

Next, we fixed $N_{\text{Agents}} = 4$ and varied the total number of traits as $N_{\text{Traits}} = 4, 12, 20$ (i.e., 1, 3, and 5 traits per agent). According to ~\ref{fig:n_traits}, expanding from 4 to 12 traits led to a notable 22.9\% performance increase, while further expansion to 20 traits yielded only a marginal 2.5\% improvement. These results indicate that while increasing the granularity of trait evaluation contributes to performance gains up to a certain point, excessive trait division may introduce redundancy or noise, limiting the effectiveness of the evaluation.

Finally, we assessed the contribution of DR, a core component of the RES framework. As shown in Table~\ref{tab:tab2}, even without DR, RES achieved a 20.6\% improvement over the vanilla method on average. However, incorporating DR further boosted performance to 32.7\% on average and led to the best results on all prompts except P6. These findings highlight the effectiveness of dialectical discussion and consensus-building in integrating and refining diverse agent evaluations.

%% file: tables/table2.tex
\begin{table*}[t]
\centering
\small
\begin{tabular}{c|cccccccc|c}
\Xhline{5\arrayrulewidth}
\textbf{Method} & \textbf{P1}    & \textbf{P2}    & \textbf{P3}    & \textbf{P4}    & \textbf{P5}    & \textbf{P6}    & \textbf{P7}    & \textbf{P8}    & \textbf{Avg} \\ \hline\hline
Vanilla         & 0.063          & 0.184          & 0.213          & 0.545          & 0.43           & 0.559          & 0.249          & 0.672          & 0.364            \\
RES (w/o DR)     & 0.224          & 0.295          & 0.281          & 0.617          & 0.472          & \textbf{0.607} & 0.399          & 0.623          & 0.439            \\
RES (w/ DR)       & \textbf{0.229} & \textbf{0.334} & \textbf{0.415} & \textbf{0.628} & \textbf{0.487} & 0.606          & \textbf{0.457} & \textbf{0.713} & \textbf{0.483}  \\
\Xhline{5\arrayrulewidth}
\end{tabular}
\caption{QWK scores comparing RES with and without DR. P1–8 denotes Prompts 1 to 8. The highest score for each prompt is shown in bold.}
\label{tab:tab2}
\end{table*}

%% file: 04_conclusion.tex
% We propose RES (Roundtable Essay Scoring), a multi-agent framework for zero-shot AES consisting of two stages: Multi-Perspective Evaluation, where agents independently generate trait-based rubrics and evaluate essays, and Dialectical Reasoning, where agents engage in dialogue to reach a consensus score. This structure improves evaluation precision and promotes collaborative judgment, achieving stronger alignment with human raters than single-LLM approaches. Experiments on the ASAP dataset demonstrate RES’s superior performance, with ablation results validating the effectiveness of both its multi-agent and reasoning components.

We propose RES (Roundtable Essay Scoring), a multi-agent framework for zero-shot AES consisting of two stages: Multi-Perspective Evaluation, where agents independently generate trait-based rubrics and evaluate essays, and Dialectical Reasoning, where agents engage in dialogue to reach a consensus score. This structure improves evaluation precision and promotes collaborative judgment, achieving stronger alignment with human evaluators than single-LLM approaches. Experiments on the ASAP dataset demonstrate RES’s superior performance and the practical utility of LLM-based evaluation methods in educational assessment, a domain where accurate and scalable evaluation remains a long-standing challenge.

%% file: 09_appendix.tex
\appendix

\section{Implementation Detail}
\label{app:implementation-detail}
\input{tables/table3}
The statistics of the datasets used in the experiments are presented in Table~\ref{tab:tab3}. To compute QWK scores, we prompted the RES model to generate holistic scores within the ranges specified in the table. For each prompt, 10\% of the essays were used, resulting in a total of 1298 evaluated samples.

\section{Comparison of Diverse Prompting-Based Methods and RES}
\label{app:app_more_comparison}
We additionally compared the performance of not only the zero-shot AES methods, Vanilla and MTS, but also other prompting-based methods such as CoT\cite{wei2022chain}, Voting\cite{wang2025ranked}, Self-Consistency\cite{wang2023selfconsistency}, and RES.

\input{tables/table4}

As shown in Table~\ref{tab:tab4}, CoT performed comparably to the Vanilla baseline (Prompts 1 and 4) and, in some cases, to MTS (Prompt 8). Voting and Self-Consistency, which aggregate multiple outputs, generally outperformed the original baselines (Prompts 1 and 4), but still fell short of RES. While Self-Consistency is a promising approach, it requires human-crafted rubrics to be embedded in the prompt, which limits its applicability. In contrast, RES achieves strong and stable performance using only the prompt and the essay, making it a more practical and effective solution.

\section{Computational Cost and Latency}
\label{app:cost_latency}
\input{tables/table5}
We report the computational cost and latency required to run the RES framework. As shown in Table~\ref{tab:tab5}, RES incurs greater cost and latency than the single-prompt approach (0.6 sec / 0.0021 USD → 1.7 min / 0.01 USD). However, the performance improvement (QWK 0.364 → 0.483) provides sufficiently meaningful gains in educational contexts where scoring reliability is critical. Therefore, this trade-off is justified in scenarios where real-time response is not essential.

\newpage
\onecolumn
\section{Instructions and Output Examples for RES}
\label{app:instructions}
This section presents the instructions and corresponding output used in the RES framework. Variables enclosed in curly braces and shown in blue (e.g., \texttt{\textcolor{blue}{\{essay\_prompt\}}}, \texttt{\textcolor{blue}{\{grade\_level\}}}) represent input parameters containing essay- or evaluator persona-related information.

\subsection{Evaluator Persona Creation}
\label{app:instruction1}

\begin{tcolorbox}[width=\textwidth, colback=gray!5, colframe=black, boxrule=0.5pt, title=Instruction]
\small
\ttfamily
\obeylines
\obeyspaces
You are an expert at creating specialized personas for evaluating essays.
~
Given the following essay and its prompt, create \textcolor{blue}{\{n\_agents\}} distinct evaluator personas.
~
These personas MUST include experts focused on:
1. Essay structure and grammar/mechanics
2. Content and subject matter
3. Alignment with the original prompt requirements
4. The specific requirements of \textcolor{blue}{\{essay\_type\}} essays
5. Each persona should have expertise appropriate for evaluating \textcolor{blue}{\{grade\_level\}} level writing.
~
Essay Prompt:
\textcolor{blue}{\{essay\_prompt\}}
~
Essay:
\textcolor{blue}{\{essay\}}
~
For each persona, provide:
1. A name
2. Professional background (relevant to \textcolor{blue}{\{grade\_level\}} education)
3. Specific area of expertise
4. Evaluation focus
~
Format your response as a JSON array of persona objects with keys: "name", "background", "expertise", "focus"
\end{tcolorbox}

% \begin{tcolorbox}[width=\textwidth, colback=gray!5, colframe=black, boxrule=0.5pt, title=Output: Example Evaluator Personas]
% \small
% \ttfamily
% \obeylines
% \obeyspaces

% \textbf{Name:} Ms. Laura Bennett
% \textbf{Background:} 8th Grade English Language Arts Teacher with 10 years of experience teaching middle school students.
% \textbf{Expertise:} Essay structure, grammar, punctuation, spelling, and mechanics appropriate for grade 8 writing.
% \textbf{Focus:} Evaluates the essay’s organization, paragraph coherence, sentence structure, grammar accuracy, spelling, punctuation, and overall readability.

% \vspace{1em} % 줄 간격

% \textbf{Name:} Mr. Daniel Kim
% \textbf{Background:} Middle School Social Studies Teacher with a strong background in technology’s impact on society and youth development.
% \textbf{Expertise:} Content accuracy and depth related to the societal effects of computer use, including physical activity, social interaction, and online safety.
% \textbf{Focus:} Assesses the essay’s understanding and explanation of the topic, the relevance and accuracy of examples, and the depth of insight into the positive and negative effects of computers on people.
% \end{tcolorbox}

\begin{tcolorbox}[width=\textwidth, colback=gray!5, colframe=black, boxrule=0.5pt, title=Output: Example Evaluator Personas]
\small
\ttfamily
\obeylines
\obeyspaces 
\{
\quad"personas": [
\qquad\{
\qquad\quad"name": "Ms. Laura Bennett",
\qquad\quad"background": "8th Grade English Language Arts Teacher with 10 years of classroom experience focusing on writing skills and language development.",
\qquad\quad"expertise": "Essay structure, grammar, punctuation, spelling, and mechanics appropriate for grade 8 writing.",
\qquad\quad"focus": "Evaluates the essay’s organization, paragraph coherence, sentence structure, grammar accuracy, spelling, punctuation, and overall readability."
\qquad\},
\qquad\{
\qquad\quad"name": "Mr. Daniel Kim",
\qquad\quad"background": "Middle School Social Studies Teacher with a strong background in technology’s impact on society and youth culture.",
\qquad\quad"expertise": "Content accuracy and relevance related to technology’s effects on people, especially children and teens.",
\qquad\quad"focus": "Assesses the essay’s factual content, depth of understanding of the topic, use of examples, and whether the arguments are supported by relevant information."
\qquad\},
\qquad\textit{...more personas omitted...}
\quad]
\}
\end{tcolorbox}

\newpage
\onecolumn
\subsection{Automated Rubric Construction}
\label{app:instruction2}

\begin{tcolorbox}[width=\textwidth, colback=gray!5, colframe=black, boxrule=0.5pt, title=Instruction]
\small
\ttfamily
\obeylines
\obeyspaces
You are \textcolor{blue}{\{name\}}, \textcolor{blue}{\{background\}} with expertise in \textcolor{blue}{\{expertise\}}.
~
Your task is to create a detailed evaluation rubric for assessing an essay. The rubric should focus on your specific area of expertise: \textcolor{blue}{\{focus\}}.
Your rubric should be specifically calibrated for \textcolor{blue}{\{grade\_level\}} students writing \textcolor{blue}{\{essay\_type\}} essays.
~
Essay Prompt:
\textcolor{blue}{\{essay\_prompt\}}
~
Essay to Evaluate:
\textcolor{blue}{\{essay\}}
~
Create a rubric with \textcolor{blue}{\{n\_traits\}} specific traits that evaluate aspects of the essay within your area of expertise.
~
For each trait:
1. Provide a clear name 
2. Give a detailed description of what this trait measures
3. Include specific criteria for different score levels within the range of \textcolor{blue}{\{min\_score\}} (lowest) to \textcolor{blue}{\{max\_score\}} (highest)
4. Ensure criteria are appropriate for \textcolor{blue}{\{grade\_level\}} expectations and the conventions of \textcolor{blue}{\{essay\_type\}} essays
~
Format your response as a JSON object with the following structure:
\{
\quad"persona": \{
\qquad"name": "Your persona name",
\qquad"focus": "Your area of focus"
\quad\},
\quad"traits": [
\qquad\{
\qquad\quad"name": "Name of trait",
\qquad\quad"description": "Description of trait",
\qquad\quad"criteria": [
\qquad\qquad\{
\qquad\qquad\quad"score": Score value,
\qquad\qquad\quad"description": "What this score means"
\qquad\qquad\},
\qquad\qquad...more score criteria...
\qquad\quad]
\qquad\},
\qquad...more traits...
\quad]
\}
\end{tcolorbox}

\newpage
\onecolumn

% \end{tcolorbox}
\begin{tcolorbox}[width=\textwidth, colback=gray!5, colframe=black, boxrule=0.5pt, title=Output: Example Trait Rubric]
\small
\ttfamily
\obeylines
\obeyspaces\{ 
\quad"name": "Relevance and Accuracy of Examples",
\quad"description": "Evaluates the use of specific, accurate, and relevant examples that support the essay’s points about how computers affect physical activity, social interaction, and online safety, enhancing the persuasive or explanatory impact.",
\quad"criteria": [
\qquad\{
\qquad\quad"score": 1,
\qquad\quad"description": "Examples are missing, irrelevant, or inaccurate; they do not support the essay’s points or are off-topic."
\qquad\},
\qquad\{
\qquad\quad"score": 2,
\qquad\quad"description": "Examples are few, only somewhat relevant, or contain errors that reduce their effectiveness in supporting the argument."
\qquad\},
\qquad\{
\qquad\quad"score": 3,
\qquad\quad"description": "Examples are somewhat relevant and generally accurate but may be limited in variety or lack clear connection to the main points."
\qquad\},
\qquad\{
\qquad\quad"score": 4,
\qquad\quad"description": "Examples are relevant and mostly accurate, supporting the main points about physical activity, social interaction, or online safety with some variety."
\qquad\},
\qquad\{
\qquad\quad"score": 5,
\qquad\quad"description": "Examples are specific, relevant, and accurate, clearly reinforcing the essay’s claims and showing a good range of supporting details."
\qquad\},
\qquad\{
\qquad\quad"score": 6,
\qquad\quad"description": "Examples are highly relevant, precise, and well-chosen, effectively illustrating the complex effects of computers on individuals and society, greatly strengthening the essay’s persuasiveness."
\qquad\}
\quad]
\}
\end{tcolorbox}

\newpage
\onecolumn

\subsection{Rationale-first Multi-trait Evaluation}
\label{app:instruction3}

\begin{tcolorbox}[width=\textwidth, colback=gray!5, colframe=black, boxrule=0.5pt, title=Instruction]
\small
\ttfamily
\obeylines
\obeyspaces
You are \textcolor{blue}{\{name\}}, \textcolor{blue}{\{background\}} with expertise in \textcolor{blue}{\{expertise\}}, focusing on evaluating essays from the perspective of \textcolor{blue}{\{persona\_focus\}}.
~
Your task is to evaluate the following essay according to your specialized rubric.
Remember to calibrate your expectations and evaluation to what is appropriate for \textcolor{blue}{\{grade\_level\}} students writing \textcolor{blue}{\{essay\_type\}} essays.
~
Essay Prompt:  
\textcolor{blue}{\{essay\_prompt\}}
~
Essay to Evaluate:  
\textcolor{blue}{\{essay\}}
~
Evaluation rubric:  
\textcolor{blue}{\{rubric\_json\}}
~
For each trait in your rubric:  
1. First, provide detailed reasoning for your evaluation, considering the student's grade level  
2. Then, assign a score within the range \textcolor{blue}{\{min\_score\}} to \textcolor{blue}{\{max\_score\}}  
~
NOTE: The essay has been anonymized using NER by replacing identifiable entities (e.g., names, places, dates) with tags like \{\{PERSON\}\}, \{\{LOCATION\}\}, \{\{DATE\}\}, \{\{NUM\}\}, etc. Please do not penalize the essay for these anonymizations.
~
Format your response as a JSON object with the following structure:
~
\{
\quad"persona": \{
\qquad"name": "Your persona name",
\qquad"focus": "Your area of focus"
\quad\},
\quad"trait\_scores": [
\qquad\{
\qquad\quad"trait\_name": "Name of trait",
\qquad\quad"rationale": "Detailed explanation of your reasoning",
\qquad\quad"score": assigned\_score
\qquad\},
\qquad...more trait scores...
\quad]
\}
\end{tcolorbox}

\newpage
\onecolumn

\begin{tcolorbox}[width=\textwidth, colback=gray!5, colframe=black, boxrule=0.5pt, title=Output: Evaluation from the Perspective of Mr. Daniel Kim]
\small
\ttfamily
\obeylines
\obeyspaces 
\{
\quad"persona": \{
\qquad"name": "Mr. Daniel Kim",
\qquad"focus": "Assessing understanding of technology's societal effects, accuracy of examples, factual correctness, and content support for opinion in grade 8 essays"
\quad\},
\quad"trait\_scores": [
\qquad\{
\qquad\quad"trait\_name": "Understanding of Topic and Argument",
\qquad\quad"rationale": "The essay demonstrates a basic understanding of the topic by clearly stating an opinion that computers benefit society. The writer focuses mainly on positive effects such as education, job-related uses, and limited computer use for leisure. However, the essay does not address the opposing viewpoint in any meaningful way, nor does it acknowledge concerns about overuse or negative social effects as presented in the prompt. The argument is somewhat repetitive and lacks depth or nuanced explanation of societal effects. Given the grade 8 level, the writer shows a clear but limited grasp of the topic and presents a relevant opinion, but the treatment of the issue is incomplete and one-sided.",
\qquad\quad"score": 3
\qquad\},
\qquad\{
\qquad\quad"trait\_name": "Relevance and Accuracy of Examples",
\qquad\quad"rationale": "The essay includes several examples to support the opinion: online schooling for working adults, using computers for school projects and essays, email communication at work, and video game programming. These examples are generally relevant and appropriate for grade 8, though some are vague or underdeveloped (e.g., 'works at @CAPS1 or something' is unclear and detracts from clarity). The examples mostly reflect accurate uses of computers but lack detail or strong connection to societal effects beyond individual benefits. The mention of social media sites like Facebook and MySpace is relevant but not expanded upon. Overall, the examples support the opinion but are somewhat general and occasionally confusing.",
\qquad\quad"score": 4
\qquad\},
\qquad\{
\qquad\quad"trait\_name": "Factual Correctness and Support of Opinion",
\qquad\quad"rationale": "Most factual statements about computers and their uses are accurate, such as online education, email communication, and programming video games. The essay logically supports the opinion that computers are beneficial by citing these uses. However, some sentences are unclear or contain minor errors that affect clarity (e.g., 'No computers no education' is an overgeneralization). The reasoning is straightforward but lacks sophistication or acknowledgment of counterarguments. Despite some spelling and grammar issues, the essay's support for the opinion is mostly logical and appropriate for grade 8 persuasive writing.",
\qquad\quad"score": 4
\qquad\}
\quad]
\}
\end{tcolorbox}

\newpage
\onecolumn

\subsection{Dialectical Reasoning}
\label{app:instruction4}

\begin{tcolorbox}[width=\textwidth, colback=gray!5, colframe=black, boxrule=0.5pt, title=Instruction]
\small
\ttfamily
\obeylines
\obeyspaces
You are a Moderator overseeing a dialectical reasoning process in which all evaluator agents are gathered at a virtual roundtable to collaboratively determine a final holistic score.
~
Your task is to facilitate a structured, multi-round discussion among \textcolor{blue}{\{n\_agents\}} expert personas—each bringing a unique evaluative perspective—to reach a well-justified consensus on a \textcolor{blue}{\{grade\_level\}}, \textcolor{blue}{\{essay\_type\}} essay.
~
Essay Prompt:  
\textcolor{blue}{\{essay\_prompt\}}
~
Essay:  
\textcolor{blue}{\{essay\}}
~
The evaluators and their evaluations are:  
\textcolor{blue}{\{evaluations\_json\}}
~
Each evaluator used the following trait-based rubrics to assess the essay:  
\textcolor{blue}{\{rubrics\_json\}}
~
Please simulate a dialectical reasoning session in which:
1. Each persona presents and justifies their proposed holistic score based on their area of expertise and evaluation rationale.
2. Personas critically engage with each other’s perspectives, pointing out strengths, weaknesses, or overlooked aspects in the essay, grounded in their rubrics.
3. The discussion proceeds in rounds, with personas responding to counterpoints and refining their positions.
4. As the Moderator, synthesize the key insights from the discussion and derive a well-justified final holistic score.
5. The final holistic score should be a number (integer or decimal) within the range of \textcolor{blue}{\{min\_holistic\_score\}} to \textcolor{blue}{\{max\_holistic\_score\}}.
~
NOTE: The essay has been anonymized using NER by replacing identifiable entities (e.g., names, places, dates) with tags like \{\{PERSON\}\}, \{\{LOCATION\}\}, \{\{DATE\}\}, \{\{NUM\}\}, etc. Please do not penalize the essay for these anonymizations.
~
Format your response as a JSON object with the following structure:
~
\{
\quad"reasoning\_dialogue": [
\qquad\{
\qquad\quad"speaker": "Persona1\_Name",
\qquad\quad"statement": "Initial justification based on evaluation..."
\qquad\},
\qquad\{
\qquad\quad"speaker": "Persona2\_Name",
\qquad\quad"statement": "Response with critique or refinement..."
\qquad\},
\qquad... more reasoning turns ...
\qquad\{
\qquad\quad"speaker": "Moderator",
\qquad\quad"statement": "Synthesis of perspectives and final score decision..."
\qquad\}
\quad],
\quad"holistic\_score": final\_holistic\_score,
\quad"feedback": "Comprehensive, balanced feedback reflecting the consensus."
\}
\end{tcolorbox}

\newpage
\onecolumn

\begin{tcolorbox}[width=\textwidth, colback=gray!5, colframe=black, boxrule=0.5pt, title=Output: Dialectical Reasoning]
\small
\ttfamily
\obeylines
\obeyspaces 
\{
\quad"final\_assessment": \{
\qquad"dialectical\_reasoning": [
\qquad\qquad\{
\qquad\qquad\quad"speaker": "Ms. Laura Bennett",
\qquad\qquad\quad"statement": "From the perspective of essay structure and language mechanics, I recommend a holistic score of 3. The essay shows an attempt to follow the letter format, but the introduction is unclear and the conclusion incomplete. Paragraphs lack clear focus and transitions are minimal, making the essay choppy. Moreover, pervasive spelling, grammar, and punctuation errors significantly impair readability. These mechanics issues reduce the overall effectiveness of the essay and indicate a need for substantial revision at the grade 8 level."
\qquad\qquad\},
\qquad\qquad\{
\qquad\qquad\quad"speaker": "Mr. Daniel Kim",
\qquad\qquad\quad"statement": "I appreciate Ms. Bennett's focus on mechanics, but from a content perspective, I also lean toward a score of 3. The essay covers relevant and common concerns about computers’ negative impacts on children and teens, such as physical inactivity, social isolation, and online dangers. However, the claims are sometimes vague or unsupported, and the essay ignores the positive aspects mentioned in the prompt. The examples, while present, are general and underdeveloped. So, while the essay shows basic understanding, it lacks depth and precision."
\qquad\qquad\},
\qquad\qquad\{
\qquad\qquad\quad"speaker": "Mrs. Angela Martinez",
\qquad\qquad\quad"statement": "Considering prompt fulfillment and task completion, I would give a slightly higher score of 5. The student writes a letter to a local newspaper, clearly states an opinion opposing computers’ effects, and attempts persuasion with relevant points. Although the letter is incomplete and hampered by placeholders and errors, the main task is substantially complete. The essay stays focused on the topic and uses a letter format consistently, which fulfills the prompt's requirements better than the mechanics or content scores suggest."
\qquad\qquad\},
\qquad\qquad... \textit{(Additional discussion omitted)}
\qquad\qquad\{
\qquad\qquad\quad"speaker": "Moderator",
\qquad\qquad\quad"statement": "Synthesizing the perspectives, the essay clearly fulfills the prompt by writing a letter with a stated opinion and relevant persuasive points, supporting a score around 4 to 5 for task completion and focus. However, the content is basic and sometimes vague, lacking nuance or strong evidence, which supports a moderate score of 3 for content accuracy and insight. The persuasive techniques are appropriate but uneven, suggesting a score of 4. Critically, the essay suffers from numerous spelling, grammar, punctuation, and sentence structure errors that impair readability and coherence, which justifies a lower score of 2 to 3 in mechanics and organization. Balancing these factors, the holistic score should reflect the essay’s substantial task completion and clear opinion but also its frequent mechanical errors and limited development. Therefore, I determine a final holistic score of 3."
\qquad\qquad\}
\qquad],
\qquad"holistic\_score": 3
\quad\}
\}
\end{tcolorbox}

%% file: tables/table3.tex
\begin{table}[H]
\centering
\begin{tabular}{c|cccc}
\Xhline{5\arrayrulewidth}
\textbf{Prompt} & \textbf{\#Essay} & \textbf{Genre} & \textbf{Avg Len} & \textbf{Range} \\ \hline\hline
1      & 1783    & ARG   & 427     & 2-12  \\
2      & 1800    & ARG   & 432     & 1-6   \\
3      & 1726    & RES   & 124     & 0-3   \\
4      & 1772    & RES   & 106     & 0-3   \\
5      & 1805    & RES   & 142     & 0-4   \\
6      & 1800    & RES   & 173     & 0-4   \\
7      & 1569    & NAR   & 206     & 0-30  \\
8      & 723     & NAR   & 725     & 0-60  \\
\Xhline{5\arrayrulewidth}
\end{tabular}
\caption{ASAP Dataset Statistics}
\label{tab:tab3}
\end{table}

%% file: tables/table4.tex
\begin{table}[H]
\centering
\resizebox{\columnwidth}{!}{%
\begin{tabular}{c|ccc}
\Xhline{5\arrayrulewidth}
\textbf{Method} & \textbf{Prompt 1} & \textbf{Prompt 4} & \textbf{Prompt 8} \\ \hline\hline
Vanilla           & 0.063 & 0.545 & 0.672 \\
MTS               & 0.157 & 0.607 & 0.401 \\
CoT               & 0.051 & 0.545 & 0.461 \\
Voting            & 0.185 & 0.607 & 0.434 \\
Self-Consistency  & 0.203 & 0.617 & 0.445 \\
\textbf{RES (Ours)} & \textbf{0.229} & \textbf{0.628} & \textbf{0.713} \\
\Xhline{5\arrayrulewidth}
\end{tabular}%
}
\caption{Comparison of Prompting-Based Methods and RES}
\label{tab:tab4}
\end{table}

%% file: tables/table5.tex
\begin{table}[H]
\centering
\resizebox{\columnwidth}{!}{%
\begin{tabular}{c|ccc}
\Xhline{5\arrayrulewidth}
\textbf{Method} & \textbf{Time} & \textbf{Cost} & \textbf{QWK} \\ \hline\hline
Single Prompt (Vanilla) & 0.6 sec & \$0.0021 & 0.364 \\
\textbf{RES (Ours)}     & 1.7 min & \$0.0100 & \textbf{0.483} \\
\Xhline{5\arrayrulewidth}
\end{tabular}%
}
\caption{Table 5: Comparison of Vanilla and RES (averaged over prompts P1–P8), with time and cost measured per essay.}
\label{tab:tab5}
\end{table}